\documentclass[10pt,twocolumn,letterpaper]{article}

\usepackage[pagenumbers]{wacv} 

\usepackage{graphicx}
\usepackage{amsmath}
\usepackage{amssymb}
\usepackage{booktabs}
\usepackage{graphicx}
\usepackage{booktabs}
\usepackage{multicol}
\usepackage{algorithm}
\usepackage{algorithmic}
\usepackage{listings}
\usepackage{dsfont}
\usepackage{color}
\usepackage{xcolor}
\usepackage{multirow}
\usepackage[]{pifont}
\usepackage{anyfontsize}

\usepackage[accsupp]{axessibility} 

\usepackage[pagebackref,breaklinks,colorlinks]{hyperref}
\usepackage{soul}

\usepackage[capitalize]{cleveref}
\crefname{section}{Sec.}{Secs.}
\Crefname{section}{Section}{Sections}
\Crefname{table}{Table}{Tables}
\crefname{table}{Tab.}{Tabs.}

\usepackage[pagebackref,breaklinks,colorlinks]{hyperref}
\def\wacvPaperID{718} 
\def\confName{WACV}
\def\confYear{2025}

\newcommand{\tick}{\ding{51}}
\newcommand{\cross}{\ding{55}}
\DeclareMathOperator{\sign}{sign}

\begin{document}

\title{Looking at Model Debiasing through the Lens of Anomaly Detection}

\author{
Vito Paolo Pastore$^{1}$\thanks{These authors contributed equally.},~~ 
Massimiliano Ciranni$^{1}$\footnotemark[1],~~
Davide Marinelli$^{1}$,~~
Francesca Odone$^{1}$,~~
Vittorio Murino$^{1,2,3}$ \\
{\normalsize $^{1}$MaLGa, DIBRIS, University of Genoa, Italy \quad 
$^{2}$Istituto Italiano di Tecnologia, Genoa, Italy \quad 
$^{3}$University of Verona, Italy}\\
}
\maketitle
\renewcommand\thefootnote{} 
\footnotetext{Correspondence: \tt\small vito.paolo.pastore@unige.it}
\footnotetext{Work accepted at \textit{IEEE/CVF Winter Conference on Applications of Computer Vision (WACV) 2025}.}
\renewcommand\thefootnote{\arabic{footnote}} 

\begin{abstract}
Deep neural networks are likely to learn unintended spurious correlations between training data and labels when dealing with biased data, potentially limiting the generalization to unseen samples not presenting the same bias. In this context, model debiasing approaches can be devised aiming at reducing the model's dependency on such unwanted correlations, either leveraging the knowledge of bias information or not.
In this work, we focus on the latter and more realistic scenario, showing the importance of accurately predicting 
the bias-conflicting and bias-aligned samples to obtain compelling performance in bias mitigation. 
On this ground, we propose to conceive the problem of model bias from an out-of-distribution perspective, introducing a new bias identification method based on \textit{anomaly detection}. 
We claim that when data is mostly biased, bias-conflicting samples can be regarded as outliers 
with respect to the bias-aligned distribution in the feature space of a biased model, thus allowing for precisely detecting them with an anomaly detection method. 
Coupling the proposed bias identification approach with bias-conflicting data upsampling and augmentation in a two-step strategy, 
we reach state-of-the-art performance on synthetic and real benchmark datasets.
Ultimately, our proposed approach shows 
that the data bias issue does not necessarily require complex debiasing methods, given that an accurate bias identification procedure is defined. Source code is available at \url{https://github.com/Malga-Vision/MoDAD}
\end{abstract}
\section{Introduction}
\label{sec:intro}
\begin{figure}
    \centering
    \resizebox{0.35\textwidth}{!}{\includegraphics{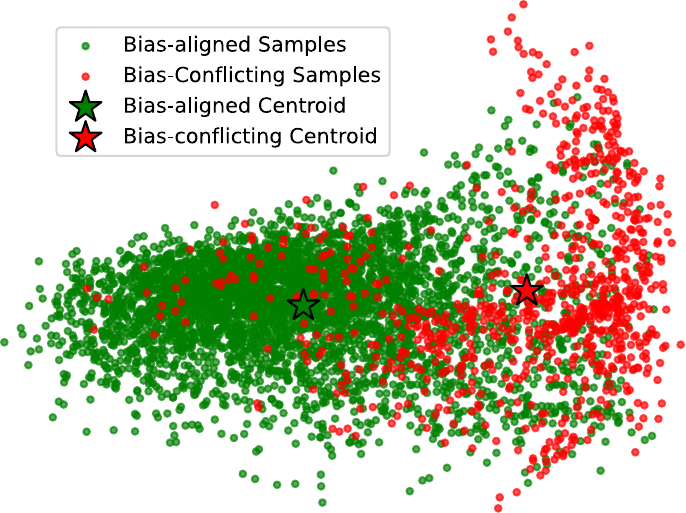}}
    \caption{First two PC visualization of bias-aligned (green dots) and bias-conflicting (red dots) ResNet-18 test features from the \textit{Cat} class in Corrupted CIFAR-10.}
    \label{fig:pca-shift}
\end{figure}

It is commonly acknowledged that deep neural networks can reach outstanding performance 
when used to tackle several computer vision tasks, including image classification, segmentation, and object detection, to name a few. Nonetheless, in case of biased data, deep neural networks often rely on spurious correlations between data and labels rather than capturing the semantic features of the objects of interest, leading to poor generalization performance. 
In this context, \textit{model debiasing} methods have been adopted to reduce the model dependency on the spurious correlations in a biased dataset, 
enhancing the model capacity to capture semantic target features, 
while mitigating the bias effect. 
As an example, let us consider the Waterbirds dataset associated with the binary classification task of predicting images of water and land birds \cite{GroupDRO}. 
In this collection, most images of water birds show water in the background, which is generally absent in images of land birds. Consequently, a deep neural network trained on such data learns an undesired shortcut, likely focusing on the background to make predictions, rather than the actual appearance of the birds. 
\\
\indent In the last few years, several methods have been proposed to deal with biased datasets. The existing works for bias mitigation either exploit ground-truth bias annotations (regarded as supervised debiasing methods) or do not assume any prior knowledge of the bias (denoted as unsupervised debiasing methods). While supervised approaches operate a finer control over the learning process allowing to selectively mitigate the bias effects, and typically leading to higher performances, unsupervised methods operate blindly, both bias attributes and biased samples are unknown, 
which is typically the case in a realistic scenario. 
\\ Among the unsupervised debiasing methods, common approaches leverage 2-stage procedures aiming at, first, detecting the samples affected by the bias (denoted as \textit{bias-aligned}) and those that are not (\textit{bias-conflicting}); and second, applying an actual debiasing process exploiting the estimated (and approximate) split of the data \cite{JTT, sohoni2020no, BiaSwap,creager2021environment, ahmed2020systematic}. Among existing debiasing works, the general assumption is that bias-aligned represents the majority of training samples (from $95\%$ to $99.5\%$ in common benchmark datasets), and we share this setting in our work.
Empirical evidence shows that the more accurate the prediction of bias-conflicting samples, the higher the generalization performances of the debiased model. 
This is key to the entire process, yet quite challenging since prior (bias) information is missing, but pervades the large majority of the data, making extremely problematic the detection of the few bias-conflicting samples. \\
\indent On this ground, we propose that bias-conflicting samples can be considered as outliers in a biased distribution of the feature space inferred by a (biased) model trained on this data. 
To support our intuition, we provide an example in Fig. \ref{fig:pca-shift}. Specifically, we train a ResNet-18 model for classification on the training set of Corrupted CIFAR-10, a synthetic biased dataset typically used as a benchmark for model debiasing. We then extract the embeddings of the training set, on which we apply Principal Component Analysis (PCA) for visualization. Then, we input the trained model with bias-conflicting and bias-aligned test samples to extract their embeddings and project them using the same Principal Component (PC) transformation applied to the training set.  
In Fig. \ref{fig:pca-shift}, we show the first two PCs of the feature space obtained in this way (green dots for bias-aligned and red dots for bias-conflicting in the figure).
For the sake of simplicity, we visualize the feature space for the \textit{Cat} class only. From this figure, one can notice that bias-conflicting samples experience a shift in the feature space distribution of the ResNet-18 model, as compared to the distribution of the bias-aligned ones. 
This simple example evidences the existence of a distribution shift between bias-conflicting and bias-aligned samples and, given the strong imbalance of such subsets, suggests the usage of an anomaly detection method. Specifically, the shift is limited to the marginal distribution of the embeddings, it is present in each class (in different forms depending on the type of bias), and it can be attributed to a covariate shift in the data \cite{nair2019covariate}. 
This can be cast as an outlier detection problem, for which anomaly detection methods can be successfully adopted to estimate test samples with covariate shifts \cite{yang2024generalized}. 
\\
\indent Hence, we propose that an accurate bias-conflicting sample detection can be obtained regarding such samples as anomalies in the embedded feature space of an intentionally biased model. This is in fact a model specifically trained to learn bias-aligned samples, discouraged from learning bias-conflicting samples, and thus amplifying the shift in distribution from the bias-aligned ones. Furthermore, we cast the bias identification as a per-class problem and exploit one anomaly detector per target class. This is a fundamental aspect of our method, as the distribution shift is amplified in the feature space of single classes, thus allowing an accurate bias-conflicting identification. 
 \\ 
To sum up, the main contribution of this work is as follows. (i) We claim that bias-conflicting samples generate anomalous feature embeddings in the feature space of a biased model. Thus, we propose the usage of an anomaly detection method for identifying bias-conflicting (and aligned) samples per target class, leveraging an intentionally biased model. To the best of our knowledge, this is the first work that proposes to look at bias-conflicting samples as anomalies, paving the way to the usage of anomaly detection methods for model debiasing tasks, and potentially opening new scenarios to bridge the gap between these two research areas.
(ii) We validate our proposed bias-identification approach coupling it with a debiasing method based on predicted bias-conflicting upsampling and data augmentation to fine-tune a model pre-trained on bias data with vanilla ERM. 
We refer to the resulting two-step method as MoDAD, Model Debiasing with Anomaly Detection. Our results show that our approach based on anomaly detection bias identification reaches competitive performances on a synthetic dataset and state-of-the-art accuracy for three real-world benchmark datasets, obtaining an average improvement of $\sim 3\%$.  
\\ \indent The remainder of the paper is organized as follows. In Section \ref{sec:relworks}, we describe relevant literature for model debiasing in image classification. In Section \ref{sec:approach}, we provide a theoretical background on anomaly detection and details of the proposed approach. Section \ref{sec:experiments} illustrates the datasets, the evaluation protocol, and implementation details, while also providing the results and ablation analyses. Finally, Section \ref{Conclusion} summarizes the proposed method and draws future work.
\section{Related Works}
\label{sec:relworks}
The literature addressing data bias and model debiasing is rich, presenting diverse approaches and methodologies that aim at reducing the impact of bias on the generalization capabilities of neural networks. Among them, supervised debiasing methods utilize explicit bias annotations, while unsupervised approaches do not rely on any prior information about the bias.  
Existing supervised debiasing methods include distributionally robust optimization \cite{GroupDRO}, and bias feature disentanglement \cite{CVPR_2021_END}. 
In ReBias, Bahng \etal~\cite{ReBias} propose an approach relying on prior knowledge on the bias rather than bias labels. Wang \etal~\cite{wang2018learning} introduce HEX, a method based on a differentiable neural building block, to reduce dependence on image textural information.
\\ \indent We focus on state-of-the-art analysis on unsupervised debiasing approaches, which share the same high-level structure of our proposed method. 
Nam \etal~\cite{LfF} propose an architecture based on two models. The first model is intentionally biased using a generalized cross-entropy loss function, while a second model is jointly trained with the first one with a weighted cross-entropy loss function, under the assumption that bias-conflicting samples would contribute more to the loss than the bias-aligned ones.
Lee \etal~\cite{DisEnt} demonstrate the importance of training a model with diverse bias-conflicting samples, proving that “diversity” outperforms oversampling. They synthesize diverse bias-conflicting samples from the bias-aligned ones.
\\ \indent
Among the unsupervised methods, a common approach involves the usage of two-step algorithms, where the first step consists in estimating the correct split between bias-conflicting and bias-aligned samples, which is further exploited for debiasing the model, typically devoting more importance to bias-conflicting samples during training. In \cite{JTT}, a model with an early stopping criterion based on explicit bias information in a validation set is first trained. The model's misclassified samples are then selected as bias-conflicting, and a second model is trained with a weighted cross-entropy loss function, whose weights are set proportionally to the number of identified bias-aligned and conflicting samples. Kim \etal~\cite{BiaSwap} provide a data augmentation framework to generate “bias-swapped” images, accounting for the bias attributes from the bias-conflicting images, and the bias-irrelevant attributes from the bias-aligned samples. Their proposed bias identification strategy consists in assigning a bias score to each sample based on both the classification correctness and model confidence, using a model trained with Generalized Cross-Entropy (GCE) loss. Seo \etal~\cite{BPA} propose a method based on feature clustering and cluster re-weighting, starting from the observation that examples with the same bias attributes tend to have similar representations in the feature space. In \cite{creager2021environment}, assuming that the training samples are partitioned into domains or environments, the authors aim to find environments that maximally violate the invariant learning principle. In these environments, the reference classifier associates the same feature vector with examples of different classes. Features that are domain-invariant predict the true class regardless of the domain. A similar bias identification approach is used in \cite{ahmed2020systematic}. In \cite{sohoni2020no}, the hidden stratification phenomenon is tackled, which happens when classes comprise multiple finer-grained sub-classes. The identification of the sub-classes is achieved through feature clustering applied on a model trained on the super-class classification task.
\\ \indent Differently from the above works, MoDAD is a two-step debiasing method  
that, first, leverages the accurate identification of 
bias-conflicting samples regarded as anomalies in the feature space of an intentionally biased model, due to a distribution shift with respect to the more numerous bias-aligned ones. Consequently, MoDAD exploits an anomaly detection algorithm for identifying the correct split of bias-conflicting and bias-aligned samples. Second, in the actual MoDAD's debiasing step, a biased model is fine-tuned using the bias-conflicting and bias-aligned sample estimates to mitigate the bias dependence.
\begin{figure*}[bt]
    \centering
    \resizebox{0.6\textwidth}{!}{
    \includegraphics{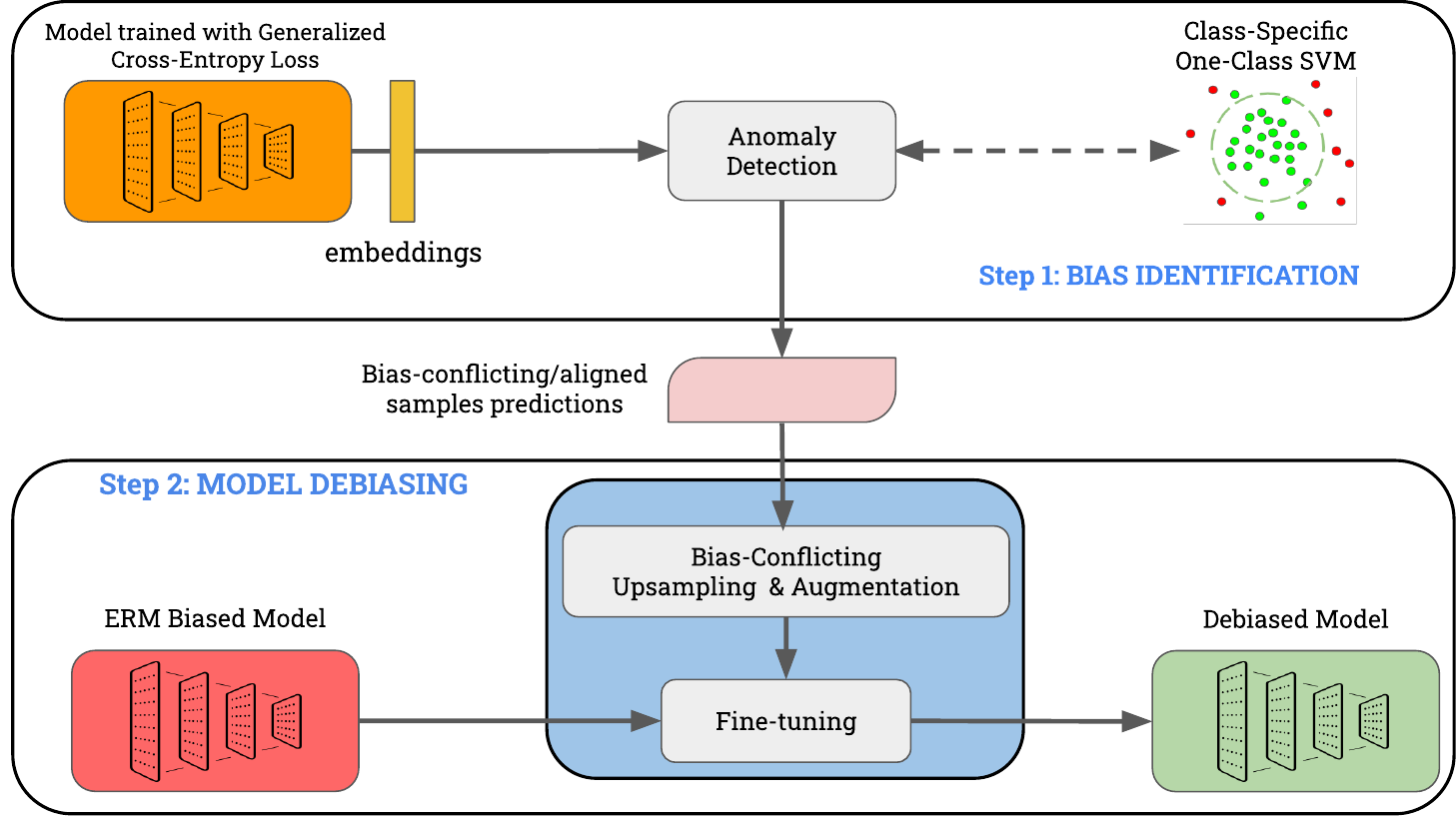}
    }
    \caption{Schematic representation of our two-step method (MoDAD).}
    \label{fig:modad-scheme}
\end{figure*}
\section{Approach}\label{sec:approach}
In this section, we describe the core part of this study. First, we detail the problem setting and outline the intuition supporting our bias-identification approach. Then, we describe the customized anomaly detection method employed in our experiments and the debiasing approach implemented for supporting the anomaly detection-based bias identification procedure.
\subsection{Problem Formulation}
Let us consider an image classification problem, where a dataset $\mathcal{D} = \lbrace \left(x_0, y_0 \right), \left(x_1, y_1 \right), \dots, \left(x_n, y_n \right) \rbrace$, for which every pair $(x_i, y_i)$ represents an input image and its associated semantic target label, is used to train a model $f_{\theta}$ to output $\hat{y}$ as the correct class of an unseen sample $x_{\text{new}}$ by minimizing an ERM objective.
If the training set $\mathcal{D}$ presents a bias
and the model is naively trained, $f_{\theta}$ likely learns shortcuts based on spurious correlations between samples and target classes instead of the actual semantic attributes \cite{LfF}. In a biased dataset, most samples present these spurious correlations (i.e., are biased-aligned), while only a small portion does not (i.e., bias conflicting), by definition. Consequently, the model naively trained on a biased dataset shows poor generalization performance when tested on new data drawn from an \textit{unbiased} distribution, as it memorizes the few bias-conflicting samples present in the training set instead of properly learning their semantics.  

In this scenario, we can expect that the bias-aligned samples of a certain training class are compact, showing high density in the model's feature space, as they share bias attributes. Intuitively, the distribution $p(X \vert Y)$ of a biased training class will be skewed, with the bias-aligned samples being concentrated in the main mode (a wide peak) and the bias-conflicting belonging to the (potentially long) tails of such distribution. This is the typical setting for Anomaly Detection, defined as the task of revealing uncharacteristic samples in a data set, defined as anomalies or outliers. Anomaly Detection methods can effectively learn a margin separating the samples in the high-density region (bias-aligned) from the samples belonging to the distribution's tails, which will fall outside of it, as they are in lower-density regions (bias-conflicting). 

Hence, we claim it is possible to identify bias-conflicting samples in a biased dataset as the samples inducing \textit{anomalous} features in a biased model. To this end, we exploit and customize a well-established anomaly detection algorithm, One-Class Support Vector Machine (OCSVM) \cite{scholkopf2001estimating}, to design a \textit{bias identification} procedure capable of identifying conflicting samples with high accuracy. The use of OCSVM stems from its effectiveness, efficiency, and controllability, while we leave to future work investigating Deep Anomaly Detection approaches for this task.

%
%
\subsection{OCSVM Customization and Bias Identificaton}\label{sec:ad-background}
\subsubsection{OCSVM customization}
Several Anomaly Detection (AD) methods have been proposed, either relying on deep learning  \cite {pang2021deep} or traditional machine learning approaches \cite{liu2008isolation,estimator1999fast,scholkopf2001estimating}.
In this work, we aim to frame the problem of unknown bias identification in the context of anomaly detection, and we opt to exploit the One-Class Support Vector Machine (OCSVM). It can be regarded as an unsupervised technique that learns a hyperplane separating the training samples from the origin while maximizing the hyperplane distance from it, and which can learn meaningful margins without the need for manual annotations \cite{scholkopf2001estimating, lamrini2018anomaly}. In the context of dataset bias, its associated decision function would ideally return $+1$ for in-class (bias-aligned) samples and $-1$ for samples detected as anomalous (bias-conflicting). 
\begin{equation}\label{eq_f}
    f(x) = \sign \left( \sum_{i=1}^m \space \alpha_i K\left(x,x_i\right) - \lambda \right)
\end{equation}
Such decision function is outlined in Eq. \eqref{eq_f}, where $\alpha_i$ and $\lambda$ are the learned Lagrange multipliers and hyperplane's offset respectively, while $K$ is the Gram matrix of distances between sample points in kernel space.
The higher the argument of the \textit{sign}, the higher the (signed) distance of a sample from the learned margin, with a positive distance indicating that a sample should be regarded as in-class (positive sign), or out-of-class (negative sign). As such, the argument of the \textit{sign}, can be regarded as an \textit{anomaly score}. In our bias identification approach, we modify Eq. \eqref{eq_f} to shift the argument of the \textit{sign} by a threshold $\tau$ (which we automatically compute, see Algorithm \ref{ps}), so that when the anomaly score is higher than $\tau$ the corresponding sample is classified as in-class, and anomalous otherwise. Such a decision threshold can be used to restrict anomaly predictions with respect to the original formulation. 
\subsubsection{Amplifying the distribution shift with GCE}
\begin{algorithm*}[tb]
    \small
    \label{pseudocode}
    \caption{Pseudocode of the proposed bias identification step}
\begin{algorithmic}[1] \label{ps}   
    \STATE \textbf{Input:} \\
    An image training set $\mathcal{D} \subseteq \mathcal{X} \times \mathcal{Y} = \lbrace (x_1, y_1), \: \dots \:, (x_N, y_N) \rbrace$, \\ 
    A Neural Network $f_{\theta}$ with randomly initialized weights $\theta$, with $f_{\theta} := \Phi \circ h$, \\
    where $\Phi$ is a DNN backbone and $h$ a NN classifier, \\
    
    \STATE \textbf{Output:} $\hat{\mathcal{B}} := \textnormal{bias-aligned and bias-conflicting samples estimation.}$ \\
    \hrulefill
    
    \STATE Define a Weighted Random Sampler $\mathcal{S}$, weighting each sample $(x_i, y_i)$ with the inverse of its class population $\psi_{y_i}$, and set \textnormal{\textit{replacement}=True} if class populations are uneven, False otherwise.
    \STATE Train $f_{\theta}$ with GCE Loss.
    \STATE Extract training samples embeddings and respective labels in a set $\mathcal{Z} = \lbrace \left(\Phi(x), y \right), \; \forall \: (x, y) \in \mathcal{D} \rbrace$.
    \STATE For each class, fit a separate $\textnormal{OCSVM}_y$ on the $\Phi(x)$ of samples from that class correctly classified by $f_{\theta}$, denoted as $\mathcal{C}_y$.
    \STATE Compute the anomaly scores of each class samples $\mathcal{A}_y$ with $\textnormal{OCSVM}_y$
    \STATE Compute a decision threshold $\tau_y$ as the anomaly score corresponding to $\textnormal{Percentile}(\mathcal{A}_y, \pi_y)$, $\pi_y = 100 \cdot \frac{1}{2} \left( \frac{\psi_y - \vert \mathcal{C}_{y} \vert}{\psi_y} \right)$    
    \STATE $\hat{\mathcal{B}_i} = \textnormal{bool}(\mathcal{A}_y\left[ \Phi(x) \right]) > \tau_y)$
    
\end{algorithmic}
\end{algorithm*}
Naively training a model on a biased dataset produces a biased classifier, learning the spurious correlations between samples and target classes, representative of bias. 
However, after just a few epochs (\cite{JTT}) the model typically overfits the training samples, memorizing bias-conflicting samples as well. 
Consequently, the actual distribution shift between bias-conflicting and bias-aligned samples in the biased model's feature space disappears. 
To overcome this issue while being robust with respect to the number of training iterations, we employ the Generalized Cross Entropy (GCE) \cite{NEURIPS2018_f2925f97_GCE} as our loss function for the bias identification step.
As shown in \cite{LfF, BiaSwap}, GCE loss makes the model to \textit{attach} onto bias, weighting the samples that the model classifies with high confidence (i.e., bias-aligned) more than the uncertain ones (i.e., bias-conflicting). However, differently from \cite{LfF, BiaSwap}, we use GCE to further amplify the distribution shift between bias-aligned and bias-conflicting samples, avoiding bias-conflicting sample memorization and favoring their identification with anomaly detection algorithms. 
\subsubsection{Identifying conflicting samples at a class level}
In our approach, we train an OCSVM per target class, allowing us to tailor bias detection specifically to individual categories. This is a core design choice of our approach, as we assume that the most dramatic shift happens inside the feature space of an individual class. As we employ GCE loss to train the biased model of our first step, we can expect that it will produce compact features for bias-aligned samples \cite{LfF, BiaSwap, DisEnt}, while spreading around bias-conflicting features and consistently misclassifying a great portion of them. In addition to mitigating the memorization effect, this also results in an amplified separation between bias-conflicting and bias-aligned features of a specific class: conflicting features will mainly overlap with features from other classes (as they are misclassified) and thus greatly differ from the \textit{normality} of its bias-aligned samples. This enables our customized class-specific anomaly detectors to better learn the distinction between samples with the same semantics but highly differing representations.
\subsection{Bias mitigation: fine-tuning the biased model}\label{sec:debias-step}
In this work, we design a debiasing scheme based on \textit{bias-conflicting upsampling and augmentation} (see Fig. \ref{fig:modad-scheme}). Starting from a model trained using the biased dataset with vanilla ERM and hence likely biased, we employ a \textit{weighted random sampler} with replacement, whose weights are set depending on the prediction $\hat{\mathcal{B}}$ of the bias identification step, as the inverse of the two estimated populations. 
With this sampler, during the debiasing training process, 
we build each mini-batch such that the ratio of bias-aligned and bias-conflicting samples 
is roughly balanced. We further perform upsampling adding three augmented images for each bias-conflicting sample in each mini-batch, so that the number of bias-conflicting samples in the batch results 4 times higher than the number of bias-aligned ones. Exploiting this batch construction scheme, a classic CE loss function is then used for debiasing a model. Differently from other two-step methods (e.g., JTT \cite{JTT}), MoDAD performs the debiasing process on an already biased model, facing a realistic scenario when bias is unknown. 
\section{Experiments}\label{sec:experiments}
In this section, we report an extensive experimental analysis, encompassing the description of the employed synthetic and realistic datasets, the evaluation protocol adopted, 
the obtained performance and comparisons, and ablation studies. Readers may refer to the Supplementary Materials for further information on implementation details, and data pre-processing. 
\subsection{Datasets}\label{sec:datasets}
\begin{figure}
    \centering
    \resizebox{0.5\columnwidth}{!}{\includegraphics{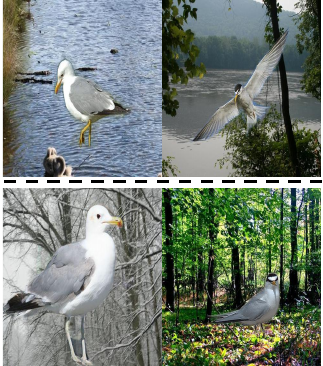}}
    \caption{Example of bias-aligned (top) and bias-conflicting samples (bottom) from Waterbirds dataset.}
    \label{fig:waterbirds-samples}
\end{figure}

To validate our approach, we consider a \textit{synthetic} (CIFAR-10) and three \textit{realistic} (BAR, BFFHQ, Waterbirds) datasets. 
The synthetic dataset allows the test of the algorithm under controlled varying conditions, specifically the knowledge of the bias and in which percentage it affects the data.  
In realistic datasets, instead, prior information regarding bias could not be available, while potentially presenting additional challenges including unbalanced per-class populations, and erroneous annotations, to name a few. 
\\
\begin{figure}[tb]
    \centering
    \resizebox{0.5\columnwidth}{!}{\includegraphics{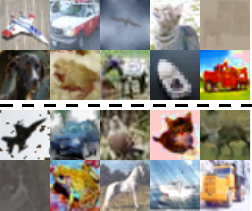}}
    \caption{Example of bias-aligned (top) and bias-conflicting samples (bottom) from Corrupted CIFAR-10 Dataset.}
    \label{fig:cifar10c-samples}
\end{figure}
\textbf{Corrupted CIFAR-10} is the first investigated dataset in the version made available from \cite{DisEnt}. This dataset is derived from the original 10-class CIFAR-10 \cite{CIFAR10Original}, composed of 60,000 images altered by 10 different types of corruptions following the protocol introduced by \cite{hendrycks2018benchmarking}. Such image corruptions are $\lbrace$ \textit{Snow}, \textit{Frost}, \textit{Fog}, \textit{Brightness}, \textit{Contrast}, \textit{Spatter}, \textit{Elastic}, \textit{JPEG}, \textit{Pixelate}, \textit{Saturate} $\rbrace$, each one being highly correlated with the ten semantic labels. We consider three degrees of correlation $\rho$ corresponding to $\lbrace 0.950, 0.980, 0.995 \rbrace$. The train and validation sets have the same correlation $\rho$ between bias-attribute and semantic label, while in the test set 90\% are bias-conflicting and only 10\% have spurious correlations with the label. Figure \ref{fig:cifar10c-samples} shows some examples from this dataset.
 \\
 \begin{figure}[tb]
    \centering
    \resizebox{0.5\columnwidth}{!}{\includegraphics{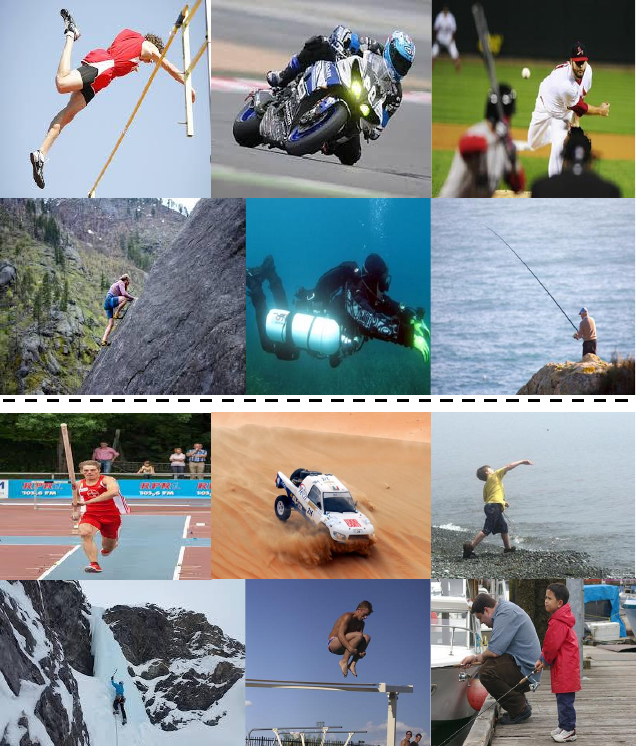}}
    \caption{Example of bias-aligned (top) and bias-conflicting samples (bottom) from BAR dataset.}
    \label{fig:bar-samples}
\end{figure}
\textbf{BAR} (Biased Action Recognition) is an action recognition image dataset with a strong correlation between the performed action and the setting in which it is carried out. It was introduced in \cite{LfF} and consists of six different semantic classes (\textit{Climbing}, \textit{Diving}, \textit{Fishing}, \textit{Racing}, \textit{Throwing}, \textit{Vaulting}), for a total of 2,595 images. The samples are further divided into a training set of 1,941 images and a test set of 654 images with no pre-defined validation set. For example, the \textit{Climbing} class contains a great majority of images in the training set where the subject is performing rock climbing. In contrast, in the test set, we can find many images of people climbing on snow or ice (see Figure \ref{fig:bar-samples}). 
Additionally, this dataset does not provide ground-truth annotation regarding whether a sample is bias-aligned or bias-conflicting.\\
\begin{figure}
    \centering
    \resizebox{0.7\columnwidth}{!}{\includegraphics{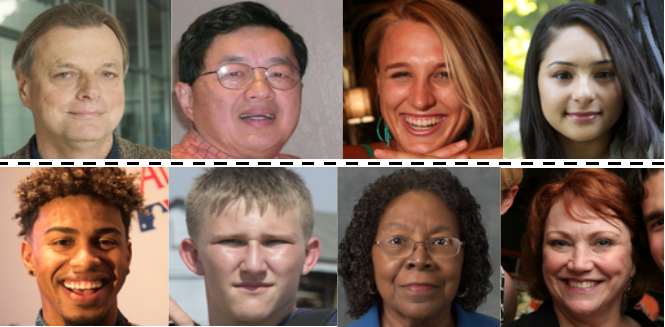}}
    \caption{Example of bias-aligned (top) and bias-conflicting samples (bottom) from BFFHQ dataset.}
    \label{fig:bffhq-samples}
\end{figure}
\textbf{BFFHQ} is a visually perceived gender-biased image dataset of human faces, obtained from the Flickr-Faces-HQ dataset \cite{karras2019style}. Introduced in \cite{BiaSwap}, we refer to the version made publicly available from \cite{DisEnt}. It is characterized by two target attributes related to age (\textit{Young} and \textit{Old}), 
where the great majority of training samples are either young female faces or old male faces (bias-aligned), while a small minority are old females or young males (bias-conflicting).
 This correlation is broken in the validation and test sets, where the two attributes are uniformly distributed. It totals 21,200 images where 19,200 images are used as training samples, 1,000 samples for validation and 1000 samples for testing. Samples of  this dataset are shown in Figure \ref{fig:bffhq-samples}.  \\ 
\textbf{Waterbirds} is an image dataset introduced in \cite{GroupDRO} that combines \textit{bird} subjects from the Caltech-UCSD Birds-200-2011 (CUB) dataset \cite{WahCUB_200_2011} and image backgrounds from the Places dataset \cite{NIPS2014_3fe94a00_places}. Images are divided according to a target attribute regarding birds being Waterbirds or Landbirds, while the background being land or water represents the bias attribute. In this setting, images depicting Landbirds on land and Waterbirds on water are bias-aligned samples, whereas pairing a Landbird with a water background constitutes a bias-conflicting sample (and vice-versa for Waterbirds on land background). Samples of this dataset are shown in Figure \ref{fig:waterbirds-samples}.
\subsection{Evaluation Protocol}\label{sec:eval-protocol}
We employ two quantitative metrics in our results: \textit{Average Accuracy} and \textit{Conflicting Accuracy}. Average Accuracy is the accuracy over the whole test set, which typically contains a balanced number of bias-aligned and conflicting samples, except otherwise stated. Conflicting Accuracy refers to the average accuracy computed only on the bias-conflicting data. We provide the mean and standard deviation of three independent runs for each experiment. For the comparative analysis, we follow the same evaluation protocol of Lee \etal~\cite{DisEnt} for Corrupted CIFAR-10, as we employ the very same version of this dataset, thus reporting the Average Accuracy of the whole test set. The test sets of BFFHQ and Waterbirds are balanced, containing an equal percentage of bias-aligned and bias-conflicting samples. For the BAR dataset, we report just the Average Accuracy due to the absence of bias annotations preventing us from computing separate metrics. Finally, as a baseline, we also report the performance of a model trained without any debiasing strategy, i.e. through Empirical Risk Minimization \cite{NIPS1991_ff4d5fbb_Vapnik} with standard cross-entropy loss.
\subsection{Results}\label{sec:results}
\textbf{Training Details} For the sake of comparison, we employ a ResNet-18 \cite{He_2016_CVPR} for Corrupted CIFAR-10, BAR, and BFFHQ, like in \cite{BiaSwap, JTT, DisEnt}. In the experiments involving Waterbirds, we opt to use a ResNet-50 as in \cite{GroupDRO}. For the real-world datasets, we start from a model whose weights are pre-trained on ImageNet \cite{ImageNet}, like in \cite{DisEnt, LfF, JTT}.  For each model, we plug on top of the last backbone layer an MLP with one hidden layer followed by a ReLU. About the OCSVM, we set the contamination parameter to $0.5$ and adopt a Radial Basis Function (RBF) kernel. For all the other technical details, please refer to the Supplementary Material.
\vspace{-1em}
\subsubsection{Corrupted CIFAR-10.}\label{sec:results-cifar} Table \ref{tab:results-comparison-cifar} provides MoDAD's results on three different $\rho$ values of Corrupted CIFAR-10, benchmarking with relevant existing approaches. Despite the basic debiasing step, our method is competitive with respect to state-of-the-art approaches for the three different bias ratios, reaching similar performances of DFA \cite{DisEnt} and LfF \cite{LfF} while being slightly lower for $\rho = 0.995$. 
Additionally, we significantly outperform JTT \cite{JTT}, whose debiasing approach is similar to the one employed in MoDAD, for all the bias ratios, further suggesting the effectiveness of our bias identification mechanism. 
Regarding the accuracy of our anomaly detection algorithm in bias identification, the average and standard deviation for the F1-score over the ten corrupted CIFAR-10 classes correspond to $0.81 \pm 0.05$, $0.80 \pm 0.07$ and $0.68 \pm 0.11$, for $\rho = \lbrace 0.950, 0.980, 0.995 \rbrace$ respectively (see Table \ref{tab:my_label}).
Finally, in the last entry of Table \ref{tab:results-comparison-cifar} we report the oracle represented by using the ground truth bias labels and our proposed debiasing method. The reported oracle accuracy provides the best performances among the referenced methods, confirming our intuition regarding the importance of precise bias identification for model debiasing. 
\begin{table}[tb]
    \centering
    \caption{F1-Score of our per-class Bias Identification method (standard deviation refers to the target classes). We provide a comparison with other works reporting this metric (or precision and recall) on the explored datasets equipped with bias annotations.}
    \label{tab:my_label}
    \addtolength{\tabcolsep}{-0.25em}
    \resizebox{\columnwidth}{!}{
    \begin{tabular}{lccccc}
        \toprule
        \textbf{Dataset}   & $\bf \rho$ &    \textbf{Ours}    & BiaSwap & GEORGE & JTT \\ \midrule
        Corrupted CIFAR-10 &  $0.950$   & $\bf 0.81 \pm 0.05$ & $0.72$ &   --   & -- \\
        Corrupted CIFAR-10 &  $0.980$   & $\bf 0.80 \pm 0.07$ & $0.69$ &   --   & -- \\
        Corrupted CIFAR-10 &  $0.995$   & $\bf 0.68 \pm 0.11$ & $0.63$ &   --   & -- \\
        Waterbirds         &  $0.950$   & $\bf 0.70 \pm 0.07$ &  --   & $0.36$ & $0.42$ \\
        BFFHQ              &  $0.995$   & $\bf 0.75 \pm 0.05$ & $0.68$ &   --   & -- \\ \bottomrule
    \end{tabular}
}

    \vspace{-1em}
\end{table}
\begin{table}[tb]
    \centering
    \caption{MoDAD's Average Accuracy on Corrupted CIFAR-10 and related comparisons. BS refers to Bias Supervision. Best results in bold, second best underlined.}
    \label{tab:results-comparison-cifar}
    \addtolength{\tabcolsep}{-0.5em}
    	\resizebox{0.475\textwidth}{!}{
		\begin{tabular}{lcccc}
			\toprule[2pt]
			\multirow{2}{*}{\textbf{Method}} &             &          \multicolumn{3}{c}{\textbf{Corrupted CIFAR-10}}           \\
			                                 & \textbf{BS} &     $\rho=0.995$     &     $\rho=0.980$     &     $\rho=0.950$     \\ \midrule
			ERM+CE                           &     --      &   $21.43 \pm 0.57$   &   $30.37 \pm 1.37$   &   $39.88 \pm 0.70$   \\ \cmidrule{3-5}
			HEX \cite{wang2018learning}      &    \tick    &  $13.87 \pm 0.06 $   &  $15.20 \pm 0.54 $   &   $16.04 \pm 0.63$   \\
			EnD \cite{CVPR_2021_END}         &    \tick    &   $22.89 \pm 0.27$   &   $31.31 \pm 0.35$   &   $40.26 \pm 0.85$   \\
			ReBias \cite{ReBias}             &    \tick    &   $22.27 \pm 0.41$   &   $31.66 \pm 0.43$   &   $43.43 \pm 0.41$   \\ \cmidrule{3-5}
			BiaSwap \cite{BiaSwap}           &   \cross    &    $\underline{29.11 \pm --}$    &    $35.25 \pm --$    &    $41.62 \pm --$    \\
			LfF \cite{LfF}                   &   \cross    &   $28.81 \pm 0.44$   &   $40.66 \pm 0.70$   &   $\underline{50.72 \pm 1.31}$   \\
			JTT \cite{JTT}                   &   \cross    &   $23.66 \pm 0.78$   &   $31.44 \pm 0.47$   &   $41.20 \pm 0.19$   \\
			DFA \cite{DisEnt}                &   \cross    & $\bf 29.95 \pm 0.71$ & $\bf 41.78 \pm 2.29$ & $\bf 51.13 \pm 1.28$ \\ \cmidrule{3-5}
			Ours                             &   \cross    &   $27.26 \pm 0.47$   &   $ \underline{41.27 \pm 0.26}$  &   $50.48 \pm 0.42$   \\ \cmidrule{3-5}
			Ours (Oracle)                    &    \tick    &    $35.98\pm0.83$    &    $45.47\pm0.83$    &    $53.84\pm0.62$    \\ \bottomrule[2pt]
		\end{tabular}                
	}    
    \vspace{-2em}
\end{table}
\begin{table*}[tb]
    \caption{MoDAD's performance on realistic image datasets. For BAR, only the average accuracy is reported, as its annotations do not contain information regarding which samples are either bias-aligned or bias-conflicting. BS refers to Bias Supervision. The best results are in bold.}
    \vspace{-1em}
    \label{tab:results-comparison-real}
    \addtolength{\tabcolsep}{-0.5em}
    \centering
    \resizebox{0.75\textwidth}{!}{
	\begin{tabular}{lcccccc}
		\toprule[2pt]
		\multirow{2}{*}{\textbf{Method}}      & \textbf{BS} &       \textbf{BAR}        &       \multicolumn{2}{c}{\textbf{Waterbirds}}       &          \multicolumn{2}{c}{\textbf{BFFHQ}}           \\
		                                      &             &     \textbf{Average}      &    \textbf{Average}     &   \textbf{Conflicting}    &     \textbf{Average}      &   \textbf{Conflicting}    \\ \midrule 
		ERM+CE                                &     --      &     $58.00 \pm 1.39$      &    $91.20 \pm 0.76$     &     $84.31 \pm 1.53$      &     $79.77 \pm 0.23$      &     $60.13 \pm 0.46$      \\
		GroupDRO \cite{GroupDRO}              &    \tick    &            --             &     $93.50 \pm --$      &     $\bf 91.40 \pm --$     &            --             &            --             \\ \cmidrule{3-7}
		BiaSwap \cite{BiaSwap}                &   \cross    &      $52.44 \pm --$       &           --            &            --             &            --             &      $58.87 \pm --$       \\
		LfF \cite{LfF}                        &   \cross    &     $62.98 \pm 2.76$      & $\mathbf{97.50 \pm --}$ &      $75.20 \pm --$       &     $75.23 \pm 1.60$      &     $62.97 \pm 3.22$      \\
		JTT \cite{JTT}                        &   \cross    &     $68.53 \pm 3.29$      &     $93.60 \pm --$      &      $86.00 \pm --$       &     $80.93 \pm 0.69$      &     $62.20 \pm 1.34$      \\
		DFA \cite{DisEnt}                     &   \cross    &            --             &           --            &            --             &            --             &     $63.87 \pm 0.31$      \\
		EIIL \cite{EIIL}                      &   \cross    &      $65.44 \pm --$       &     $96.90 \pm --$      &      $78.70 \pm --$       &            --             &     $59.20 \pm 1.90$      \\
		George \cite{GEORGE}                  &   \cross    &            --             &    $95.50 \pm 2.00$     &      $76.20 \pm --$       &            --             &            --             \\
		DebiAN \cite{li2022discover_Debian}   &   \cross    & $\mathbf{69.88 \pm 2.92}$ &           --            &            --             &            --             &     $62.80 \pm 0.60$      \\
		SIFER \cite{tiwari2023overcoming}        &   \cross    &     $65.75 \pm 1.84$      &           --            &            --             &            --             &            --             \\ 
            uLA \cite{tsirigotis2024group} & \cross & -- & $91.50 \pm 0.70$ & $86.10 \pm 1.50$ & -- & -- \\
            HSSD \cite{arora2024hybrid} & \cross & -- & -- & -- & -- & 63.86 ± 1.03 \\
\midrule		
  Ours                                  &   \cross    & $\mathbf{69.83 \pm 0.72}$ &    $93.52 \pm 0.73$     & $\mathbf{89.43 \pm 1.69}$ & $\mathbf{83.50 \pm 1.39}$ & $\mathbf{68.33 \pm 2.89}$ \\ \midrule
		Ablation w\textbackslash \;JTT        &   \cross    &     $63.25 \pm 0.38$      &    $93.00 \pm 0.09$     &     $87.93 \pm 0.13$      &     $80.20 \pm 0.02$      &     $61.13 \pm 0.12$      \\
		Ablation w\textbackslash \;Global OCSVM &   \cross    &     $61.16 \pm 2.59$      &    $92.09 \pm 0.43$     &     $86.02 \pm 1.12$      &     $79.45 \pm 1.63$      &     $59.70 \pm 3.25$      \\ \bottomrule[2pt]
	\end{tabular}
}

    \vspace{-1em}
\end{table*}
\subsubsection{Realistic Datasets}\label{sec:results-real}
Table \ref{tab:results-comparison-real} reports the performance of MoDAD and a comparison with relevant existing works on the three real-world datasets. First, we evaluate the accuracy of our bias-identification approaches on Waterbirds and BFFHQ, obtaining an average f1-score of $0.70$ and $0.75$, respectively (see Table \ref{tab:my_label}). Regarding the debiasing performance, MoDAD surpasses other unsupervised debiasing works, apart from the Average Accuracy on Waterbirds, where LfF \cite{LfF} reports the highest among all the works in this category. At the same time, our method reaches higher Conflicting Accuracy, almost matching the supervised state-of-the-art on this dataset (GroupDRO \cite{GroupDRO}). Regarding BAR, we match the performance of DebiAN \cite{li2022discover_Debian} ($-0.05\%$) on average, but with a significantly lower standard deviation ($2.92$ for DebiAN, $0.72$ for MoDAD). BFFHQ has a bias-correlation $\rho=0.995$, thus capturing the contribution of the few bias-conflicting training is much harder. However, in BFFHQ we report the most significant improvement over existing works, with a Conflicting Accuracy of $68.33\%$, corresponding to an improvement of $4.46\%$ over the second-best approach (DFA \cite{DisEnt}). 
\subsection{Ablation Studies}\label{sec:ablation}
\begin{table}[tb]
    \centering
    \caption{MoDAD’s performance on Corrupted CIFAR10 ($\rho=0.950$) with different anomaly detection algorithms}
    \label{tab:ablation-adec}
    \addtolength{\tabcolsep}{-0.25em}
    \resizebox{0.5\columnwidth}{!}{\begin{tabular}{lc}
    \toprule[2pt]
    AD Algorithm &     Average Accuracy     \\ \midrule[2pt]
    COV          &      $44.34 \pm 0.29$    \\
    IFO          &     $44.95 \pm 0.61$     \\
    LOF          &    $49.10 \pm 0.13$      \\
    OCSVM        & $\bf 50.48 \pm 0.62$ \\ \bottomrule[2pt]
\end{tabular}}  
    \vspace{-2em}
\end{table}
In this section, we report ablation studies to evaluate the impact of our design choices on the proposed methodology. Additional analyses can be found in the Supplementary.\\
\noindent \textbf{Bias identification method.} To further support the effectiveness of our anomaly detection-based bias identification procedure, we perform an ablation study on our bias identification method. Specifically, we run our debiasing step using bias-aligned and bias-conflicting predictions obtained with the original implementation of JTT \cite{JTT}. The last entry of Table \ref{tab:results-comparison-real} summarizes the obtained results. On three runs, we obtain an average decrease in the debiased model performance corresponding to  $1.63\%$ and $7.20\%$ for the Conflicting accuracy in Waterbirds and BFFHQ, respectively, and $6.58\%$ in Average accuracy for BAR. The obtained results further support the impact of the anomaly detection method for identifying bias in the performance of the final debiased model.\\
\textbf{Anomaly detection algorithm.} To show the soundness of the proposed approach regardless of the anomaly detection method adopted, we perform several tests by using other readily available anomaly detectors, on the Corrupted CIFAR-10 dataset (with $\rho = 0.950$).
Specifically, we consider the \textit{Local Outlier Factor} (LOF) \cite{lof}, the \textit{Isolation Forest} (IFO) \cite{ifo}, and the \textit{Robust Covariance Estimator} (COV) \cite{ROBUST}. Table \ref{tab:ablation-adec} reports the average accuracy on Corrupted CIFAR-10 with $\rho = 0.950$. OCSVM is indeed the best option among the considered ones. However, the obtained results confirm that changing the detector has only a modest impact on the results, demonstrating that the intuition regarding bias-conflicting samples as anomalies does not indeed depend on our specific design choice of the anomaly detection method.
\section{Conclusions}\label{Conclusion}
Deep neural networks typically exhibit poor generalization performances when bias is present in the data. In these cases, they likely rely on misleading correlations between bias and labels rather than the desired correlations representing the semantics of a certain class. Model debiasing consists in methods for mitigating the model's prediction dependency on bias. In this work, we consider an unsupervised debiasing framework, meaning that no knowledge of bias is used, simulating a realistic scenario. We claim that bias-conflicting samples produce features that are anomalous in the feature space of an intentionally biased model. Thus, we propose to use anomaly detection methods for bias identification, filling the gap between model debiasing and anomaly detection, and potentially opening new research directions. To validate the proposed strategy, we embed it into a two-step method named MoDAD, where the bias-aligned and bias-conflicting samples, identified through the anomaly detection scheme, are used for model debiasing. Specifically, our model debiasing consists in fine-tuning a biased model (trained with vanilla ERM) exploiting bias-conflicting upsampling and augmentation. 
Our results show that MoDAD can reach competitive performance on benchmark datasets, slightly outperforming state-of-the-art methods based on more complicated debiasing approaches, empirically proving the effectiveness of the anomaly detection bias identification strategy. 
{\small
\vspace{-1em}
\paragraph{Acknowledgments}
\noindent We acknowledge the financial support from PNRR MUR Project PE0000013 "FAIR", funded by the European Union - NextGeneration EU, CUP
J33C24000430007
}
{\small
\bibliographystyle{ieee_fullname}
\bibliography{egbib}
}

\clearpage
\appendix
\twocolumn[ 
    \begin{center}
        {\Large \bfseries Appendix}
    \end{center}
    \vspace{5em}
]

\section*{S1. Additional Ablation Studies} 
\subsection*{S1.1 Contribution of the per-class custom threshold $\tau_y$}
In our bias identification step, we customize the OCSVM algorithm, manipulating its decision function to accommodate a custom per-class decision threshold $\tau_y$. These thresholds (different for each class) push the anomaly detectors to reject samples as out-of-class (\ie., bias-conflicting) only when they are highly confident, and thus better discriminating between bias-aligned and bias-conflicting samples.
In Table \ref{tab:threshold-no-threshold},  we report an ablation study on Corrupted CIFAR-10 (for three considered bias correlation values $\rho$), evaluating the impact of using our custom thresholds or exploiting the original formulation with the default threshold (which is $0$). 
\begin{table}[h!]
\renewcommand\thetable{S.1} 
    \centering
    \resizebox{\columnwidth}{!}{
    \begin{tabular}{lccc}
        \toprule[2pt]
        {\textbf{Corrupted CIFAR-10}}     &   $\rho=0.995$   &   $\rho=0.980$   &   $\rho=0.950$   \\ \midrule
        \begin{minipage}{0.33\columnwidth}Default \newline threshold\end{minipage} & $23.21 \pm 0.07$ & $30.80 \pm 0.17$ & $37.88 \pm 0.35$ \\ \midrule
        \begin{minipage}{0.33\columnwidth}Per-class custom \newline threshold $\tau_y$(Ours)\end{minipage} & $\bf 27.26 \pm 0.47$ & $\bf 41.27 \pm 0.26$ & $\bf 50.48 \pm 0.42$ \\ \bottomrule[2pt]
    \end{tabular}			
}
    \caption{Average accuracies and standard deviations (over three runs) on Corrupted CIFAR-10, where we ablate the usage of custom decision thresholds in the OCSVM anomaly detectors employed in the proposed bias identification step.}
    \label{tab:threshold-no-threshold}
\end{table}

Although the default threshold can still provide reasonable debiasing performance, the proposed per-class custom thresholds lead to a significant improvement, ranging from about $+4\%$ to $13\%$, at the different bias levels.
These results support our intuition on the necessity of obtaining a \textit{purer} bias-conflicting prediction for improving debiasing performance. 
%
%
\subsection*{S1.2 Sensitivity to different anomaly detection algorithms}
\begin{figure*}[h!]
\renewcommand\thefigure{S.1} 
  \begin{subfigure}{0.49\linewidth}
    \centering
    \resizebox{\textwidth}{!}{
        \includegraphics{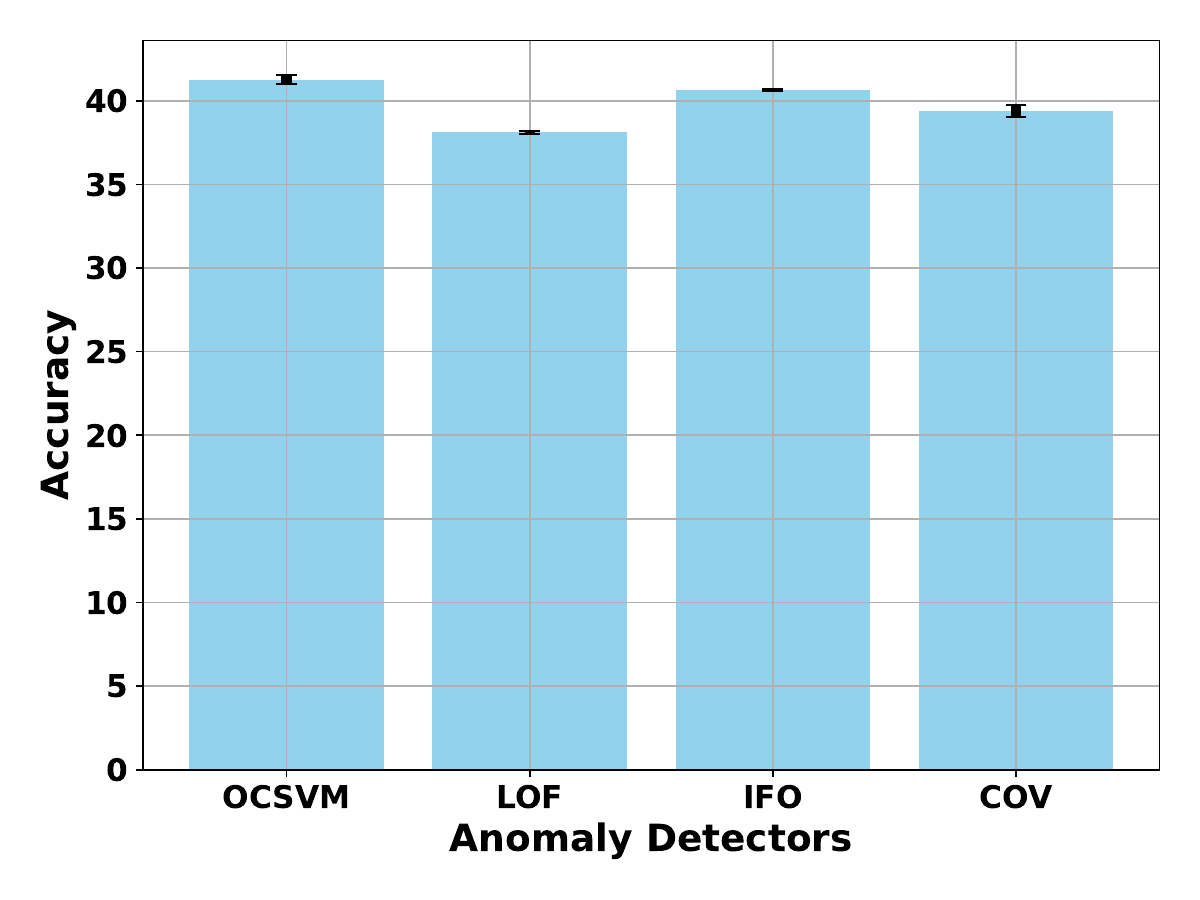}
    }
    \caption{$\rho=0.980$.}
    \label{fig:other-adecs-98}
  \end{subfigure}
  \hfill
  \begin{subfigure}{0.49\linewidth}
    \centering
    \resizebox{\textwidth}{!}{
        \includegraphics{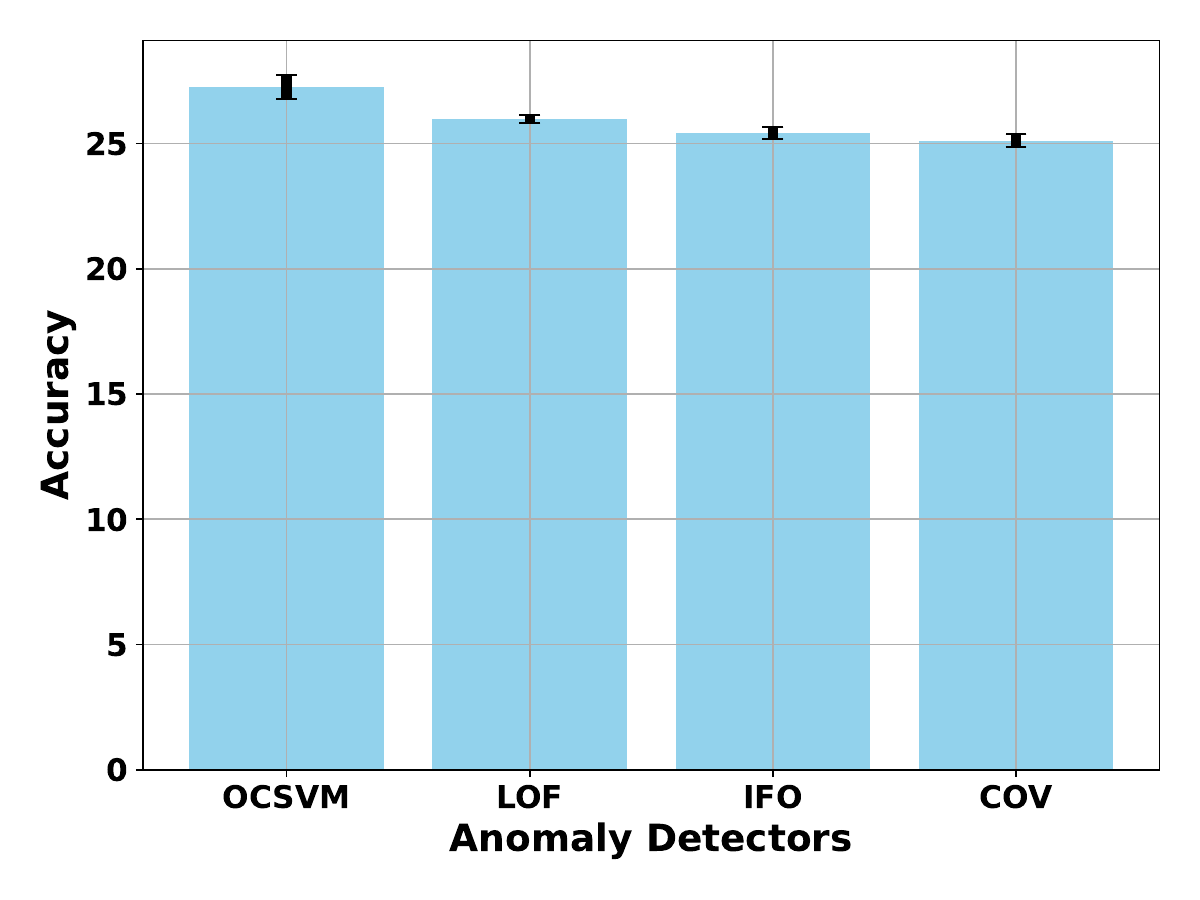}
    }
    \caption{$\rho=0.995$.}
    \label{fig:other-adecs-98}
  \end{subfigure}
  \caption{MoDAD average accuracy over three different runs with different anomaly detection algorithms on Corrupted CIFAR-10.}
  \label{fig:AD}
\end{figure*}

In Figure \ref{fig:AD}, we report the impact on Corrupted CIFAR-10 in terms of average accuracy of using different anomaly detection (AD) algorithms for bias correlation values $\rho = 0.980$ and $\rho = 0.995$ (in addition to the $\rho = 0.950$ case already provided in Section $4.4$ of the main paper). The comparison involves the same algorithms indicated in the main paper, i.e. OCSVM, \textit{Local Outlier Factor} \cite{lof}, \textit{Isolation Forest} \cite{ifo}, and \textit{Robust Covariance Estimator} \cite{ROBUST}.

As already noticed for the $\rho = 0.950$ case, even with a higher bias correlation, there are no significant differences in performance among the several AD methods. We notice only a slight decrease in the performance for the explored detectors, which remains in the range of 2-3\% with respect to the OCSVM. This empirically proves that even when the bias correlation level increases, different anomaly detectors behave similarly.
\subsection*{S1.3 MoDAD impact in case of non-biased datasets}
\begin{table}[tb]
\renewcommand\thetable{S.2} 
    \centering
    
\begin{tabular}{lc}
    \toprule[2pt]
    {\textbf{non-biased CIFAR-10}}     &  Accuracy   \\ \midrule
    ERM+CE               & $84.98 \pm 0.35$ \\
    MoDAD & $82.87 \pm 0.47$ \\ \bottomrule[2pt]
\end{tabular}			

    \caption{Impact of applying MoDAD on a model trained with the original version of CIFAR-10. Reported means and standard deviations are computed from three independent runs.}
    \label{tab:my_label}
\end{table}
Our method (MoDAD) is designed to be applied to a biased model, \ie., on a model that fails to generalize even when properly trained and tuned, suggesting the presence of bias in data or dramatic distribution shifts. 
In such a case, MoDAD has a significant impact, succeeding in reducing the bias effect and debiasing the model.  
However, in general, we do not know whether a dataset is biased or not. Hence, we are interested in assessing the effect of MoDAD when applied to an unbiased model, \ie., a model trained on an unbiased dataset. Thus, we run the two steps of MoDAD on the original non-biased version of CIFAR-10 and estimate average accuracy on the test set. As we can see in Table \ref{tab:my_label}, our debiasing method affects only slightly the performance on the test set with respect to the baseline ERM+CE trained model: we experience just a drop of $\sim 2\%$, which is significantly lower than the average benefit that we can get when using MoDAD on actually biased datasets.
\section*{S1.4 Input Model for the Debiasing Step}
\begin{table}[tb]
\renewcommand\thetable{S.3} 
    \centering    
    \begin{tabular}{lc}
    \toprule[2pt]
    Input Model   &     Average Accuracy     \\ \midrule[2pt]
    GCE Model     &     $38.35 \pm 0.35$     \\
    Biased ERM    & $\bf 50.48 \pm \pm 0.62$ \\ \bottomrule[2pt]
\end{tabular}    
    \caption{MoDAD's performance on Corrupted CIFAR10 ($\rho=0.950$) with different input models for the debiasing step.}
    \label{tab:ablation-input}
\end{table}
As reported in Section 3.3 of the main paper, our debiasing approach consists in fine-tuning a model trained with a vanilla ERM on the biased datasets. However, an alternative can be fine-tuning the intentionally biased model (trained with the GCE) utilized in the first step of our proposed approach.
In Table \ref{tab:ablation-input} we report a comparison between the two different input models for the debiasing step, on the Corrupted CIFAR-10 dataset (with $\rho = 0.950$). Fine-tuning the GCE model corresponds to a drop of $\sim 12\%$. This is expected, as the GCE model is intentionally and extremely biased by design, so it is more challenging to mitigate its bias dependency with respect to the biased Vanilla ERM model.
\section*{S2.   Implementation Details}
\label{sec:impl-details}
The code supporting this research and its experiments is implemented in \texttt{Python} \cite{Python3}, with the support of \texttt{Pytorch} and \texttt{Torchvision} \cite{NEURIPS2019_bdbca288_pytorch, torchvision2016} for data pre-processing, neural network implementation, training, and evaluation. For the OCSVM algorithm (see Section 3 of the main paper), we rely on the open-source implementation available from \texttt{Scikit-Learn} \cite{scikit-learn}. 
Additionally, generic numerical operations and data visualization are performed exploiting \texttt{Numpy} \cite{harris2020array_numpy} and \texttt{Matplotlib} \cite{Hunter:2007_matplotlib} respectively.

\subsection*{S2.1  Data Pre-Processing}
Input images are square-resized to a pixel resolution of 224x224 for BAR, Waterbirds, and BFFHQ. Corrupted CIFAR-10 images are instead kept at their original resolution of 32x32 pixels, following \cite{LfF, DisEnt, BiaSwap}. Regardless of the dataset, images are normalized to have RGB values between 0 and 1. Additionally, we replicate the same augmentations on the training images of Corrupted CIFAR-10, BAR, and BFFHQ that are found in \cite{LfF, DisEnt, BiaSwap}, i.e.: 
\begin{itemize}
    \item \textbf{Corrupted CIFAR-10}
    \begin{enumerate}
        \item \texttt{RandomCropping((32, 32))}
        \item \texttt{Padding((4, 4))}
        \item \texttt{RandomHorizontalFlip(p=0.5)}
    \end{enumerate}
    \item \textbf{BAR}
    \begin{enumerate}
        \item \texttt{RandomResizedCrop((224, 224))}
        \item \texttt{RandomHorizontalFlip(p=0.5)}
        \item \texttt{ImageNet Standardization}
    \end{enumerate}    
    \item \textbf{BFFHQ}
    \begin{enumerate}
        \item \texttt{Resize((224, 224))}
        \item \texttt{Padding((4, 4))}
        \item \texttt{ImageNet Standardization}
    \end{enumerate}
\end{itemize}
During testing, we apply only square resizing and ImageNet standardization for BAR and BFFHQ. 
Concerning Waterbirds, we only perform square resizing of all input images with a target resolution of 224x224 pixels, followed by ImageNet standardization.
We rely on \texttt{Torchvision} \cite{torchvision2016} for all the mentioned pre-processing operations.

\subsection*{S2.2 Training Details}\label{sec:additional-training-details}
In every experiment, we adopt AdamW as optimizer \cite{loshchilov2018decoupled}, with an initial learning rate of $10^{-5}$ for the debiasing step. For the bias-identification step involving GCE pre-training, we use an initial learning rate of $10^{-3}$ for Corrupted CIFAR-10 and $10^{-5}$ for the other datasets. The ERM models trained with Cross-Entropy loss (ERM+CE), are trained with a fixed learning rate of $10^{-3}$ and a mini-batch random sampler weighted on class populations. 
Regarding batch sizes, for the sake of fair comparisons, we follow what is found in the existing literature \cite{LfF, DisEnt, GroupDRO}: 256 for Corrupted CIFAR-10 and BAR, 128 for Waterbirds, and 64 for BFFHQ. \\
\noindent \textbf{Network Embeddings. } To extract the network embeddings employed in the proposed bias identification step (see Sec. 3.2 of the main paper), we consider the very last layer before the \textit{softmax} layer that performs the final classification. This corresponds to the additional layer we attach to the ResNet backbone (see Training Details in Sec. 4.3 of the main paper), which is a \textit{linear} layer with 128 neurons put after the ResNet's last \textit{pooling} layer, followed by a ReLU non-linearity. This is fixed regardless of the dataset and the backbone being a ResNet-18 (Corrupted CIFAR-10, BAR, BFFHQ) or a ResNet-50 (Waterbirds).\\
\noindent \textbf{Training Iterations and Regularization. }
The bias-identification model trained with GCE loss is trained for 100 epochs in the case of Corrupted CIFAR-10, 30 epochs for BAR and BFFHQ, and 50 epochs for Waterbirds.
The Debiasing step is performed for 100 epochs in the experiments for Corrupted CIFAR-10, while 50 epochs are employed for BAR, Waterbirds, and BFFHQ. We set a fixed number of epochs and do not employ any implicit regularization technique during training, as we cannot assume datasets-wise accordance regarding distribution shifts in the validation set. 


%
%


\end{document}